\documentclass{article}
\usepackage{spconf,graphicx}
\usepackage{microtype}
\usepackage{cite}
\usepackage{amsmath,amssymb,amsfonts}
\usepackage{algorithmic}
\usepackage{textcomp}
\usepackage{adjustbox}
\usepackage{threeparttable}
\usepackage{xcolor}
\usepackage[inline]{enumitem} 
\usepackage[symbol]{footmisc}
\usepackage{hyperref}
\hypersetup{
    colorlinks=true,
    citecolor=blue,
    linkcolor=blue,
    filecolor=magenta,      
    urlcolor=blue,
}

\title{Real-time Speech Interruption Analysis: From Cloud to Client Deployment}
\name{Quchen Fu$^{\dagger}$, Szu-Wei Fu$^{\star}$, Yaran Fan$^{\star}$, Yu Wu$^{\star}$, Zhuo Chen$^{\star}$, Jayant Gupchup$^{\star}$$^1$, Ross Cutler$^{\star}$}

\address{$^{\dagger}$ Dept. of Computer Science, Vanderbilt University \\ $^{\star}$ Microsoft Corporation}

\begin{document}
\maketitle    
\begin{abstract}
Meetings are an essential form of communication for all types of organizations, and remote collaboration systems have been much more widely used since the COVID-19 pandemic. One major issue with remote meetings is that it is challenging for remote participants to interrupt and speak. We have recently developed the first speech interruption analysis model \textit{WavLM\_SI}, which detects failed speech interruptions, shows very promising performance, and is being deployed in the cloud. To deliver this feature in a more cost-efficient and environment-friendly way, we reduced the model complexity and size to ship the WavLM\_SI model in client devices. In this paper, we first describe how we successfully improved the True Positive Rate (TPR) at a 1\% False Positive Rate (FPR) from 50.9\% to 68.3\% for the failed speech interruption detection model by training on a larger dataset and fine-tuning. We then shrank the model size from 222.7 MB to 9.3 MB with an acceptable loss in accuracy and reduced the complexity from 31.2 GMACS (Giga Multiply-Accumulate Operations per Second) to 4.3 GMACS. We also estimated the environmental impact of the complexity reduction, which can be used as a general guideline for large Transformer-based models, and thus make those models more accessible with less computation overhead.
\end{abstract}


\section{Introduction}
The inability of virtual meeting participants to interrupt and speak has been identified as the largest impediment to more inclusive online meetings \cite{cutler2021meeting}. Remote participants may often attempt to join a discussion but can not get the floor as the other speakers are talking. We refer to such attempts as \textit{failed interruptions}. 
\footnotetext[1]{Work performed while at Microsoft, now affiliated with Uber technologies.}

Too many failed interruptions may alienate participants and result in a less effective and inclusive meeting environment, and further impact the overall working environment and employee retention at organizations~\cite{fu2022improving}. Our previous work~\cite{fu2022improving} has explored the feasibility of mitigating this issue by creating a failed interruption detection model that prompts the failed interrupter to raise their virtual hands and gain attention, a feature that is helpful to improve meeting inclusiveness but rarely used.

The WavLM-based Speech Interruption Detection model (WavLM\_SI)~\cite{fu2022improving} was deployed in the Azure cloud and has proven to be a useful feature. Client integration allows us to reach a much wider range of customers without the cost of cloud deployment, as well as lower environmental impact since it does not have the overhead of a dedicated service. However, the original model is based on a pre-trained speech model, which is large and computationally expensive, and therefore can not be run on client devices. To be deployed in the client, the model needs to be small in size to meet memory constraints, computationally lightweight to incorporate less capable processors, and energy efficient to save battery life. A demo video of WaveML\_SI is available \href{https://rosscutler.github.io/speech_interruption.mp4}{here}.

This paper provides three contributions to the study of speech interruption analysis.
First, we increased the state-of-the-art performance of the speech interruption detection task from 50.9\% to 68.3\% TPR at a fixed 1\% FPR.
Second, we conducted comprehensive work on model structure engineering, including the bottleneck analysis for client integration, and created a customized model under strict computation and memory constraints. 
Third, we showed how pruning and quantizing can be used to reduce the model size by $23 \times$ and the complexity by $9 \times$ with acceptable performance, and this size/complexity and accuracy trade-off is smooth and can accommodate a large variety of client devices.

The remainder of this paper is organized as follows: Section \ref{related.section} summarizes prior work on deep learning model size reduction and energy measurement. Section \ref{modelstruct.section} introduces our base model structure and discusses structural exploration and quantization. Section \ref{experiment.section} describes our experiment setup and analyses memory usage, energy consumption, and potential environmental implications.  Section \ref{conclusion.section} provides concluding remarks and future work.

\vspace{-0.2in}
\section{Related Work}
\label{related.section}

Improving meeting inclusiveness has significant financial incentives such as more effective meetings and higher employee retention rates for organizations~\cite{cutler2021meeting}. Speech interruption analysis~\cite{fu2022improving} can help address this top issue by empowering participants to interrupt and speak in virtual meetings. 

\textbf{Self-supervised learning:}
Learnable audio frontends are gradually replacing traditional fixed frontends (e.g., spectrograms) in audio tasks~\cite{9746722}. Self-supervised learning (SSL) speech models~\cite{hsu2021hubert,baevski2020wav2vec,chen2022wavlm} have achieved state-of-the-art on a variety of tasks and dominated the leader-board of SUPERB~\cite{yang2021superb}, a performance benchmark of shared models across a wide range of speech processing tasks with minimal architecture changes and labeled data. WavLM~\cite{chen2022wavlm} is currently the best SSL speech model on the SUPERB leaderboard.

\textbf{Model size reduction:}
Our previous work~\cite{fu2022improving} utilized WavLM as the embedding layer for the speech interruption detection task and reached a TPR of 50.9\% at an FPR of 1\%. The model is 1.2 GB in size and has an inference time of 2.5 seconds processing 5 seconds of audio when running in a single thread on an Intel Xeon E5-2673 CPU (2.40GHz), which is equivalent to a Real-Time Factor (RTF) of 0.5. To reduce the model size for better inference efficiency, previous work has explored Structural Change~\cite{lee2022fithubert}, Teacher-Student Knowledge Distillation~\cite{CSS_with_TSTransformer,9747490,Wang2021MiniLMv2MS}, Parameter Sharing~\cite{ge2022edgeformer}, Layer Drop~\cite{DBLP:conf/iclr/FanGJ20}, Post Training Quantization (PTQ)~\cite{yao2022zeroquant}, and a combination of the above~\cite{wang2022lighthubert}. The most intuitive and efficient method for reducing the model size we found has been model pruning~\cite{hoefler2021sparsity}, namely only keeping the core components and using them effectively. Our target is to shrink the model to 10 MB with an RTF of 0.1 while retaining a TPR above 40\%. This target is set so that we can catch at least two failed interruptions per meeting on average, assuming 40 interruptions per half-hour meeting~\cite{fu2022improving}. 

\textbf{Energy measurement:}
Inference is estimated to account for the vast majority of the energy cost compared to training ~\cite{desislavov2021compute}. The energy associated with it, therefore, is of major concern, especially if the inference runs as a background service and has a user base size of 300 million. To encourage less energy consumption, Henderson~\cite{henderson2020towards} proposed a framework for tracking environmental impact. A specialized energy consumption study is also done for deep learning models in NLP~\cite{strubell2019energy}.

\section{Model and Method}
\label{modelstruct.section}

\subsection{Baseline Model}

\begin{figure}[hpbt]
\centering
\includegraphics[width=1.0\linewidth]{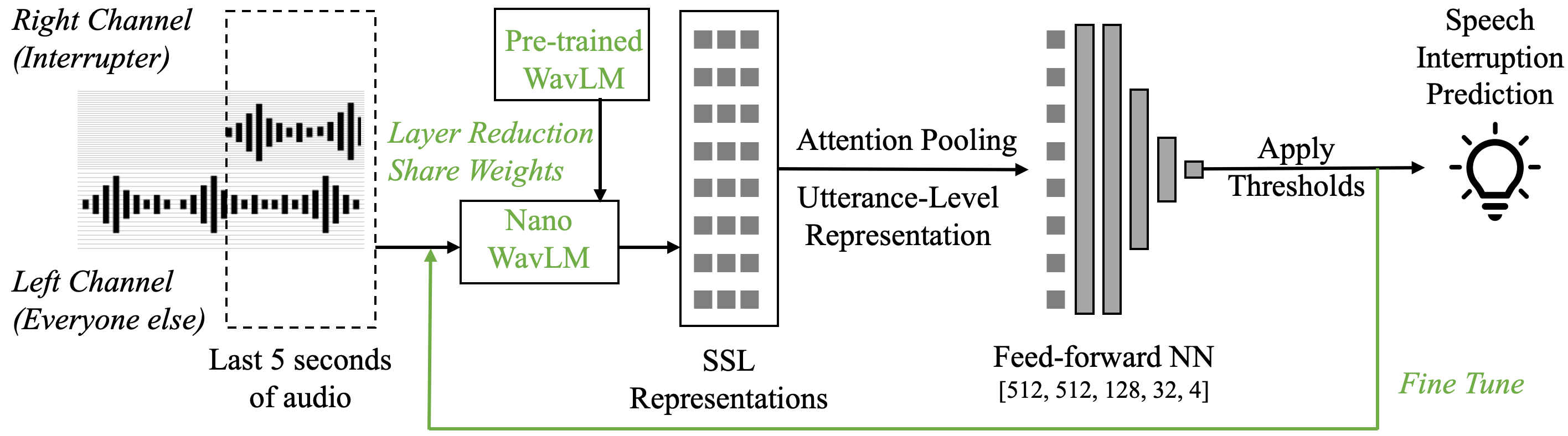}
\caption{Failed speech interruption model (WavLM-SI) architecture and training pipeline.}
\label{fig:modelstructure}
\end{figure}

Our failed speech interruption model architecture is shown in Figure~\ref{fig:modelstructure}, which consists of a WavLM-based embedding module followed by an attention pooling~\cite{tseng2021utilizing}, and a DNN-based classifier. The architecture colored in black is from our previous work and the green components are added in this paper. WavLM was pre-trained on large-scale in-house audio data and acted as a feature extractor. We multiplied all layers' output with learnable weights to get a weighted sum representation and the E2E model was fine-tuned on our dataset for the downstream task. By fine-tuning, together with a better and larger dataset, we improved the baseline model TPR at 1\% FPR from 50.9\% to 68.3\% with a much smaller model, 
and reduced the RTF from 0.5 to 0.32.

\subsection{Model Compression}
Here we discuss our guidelines for model pruning and strategies for weight sharing.

\begin{figure}[hpbt]
\centering
\includegraphics[width=1.0\linewidth]{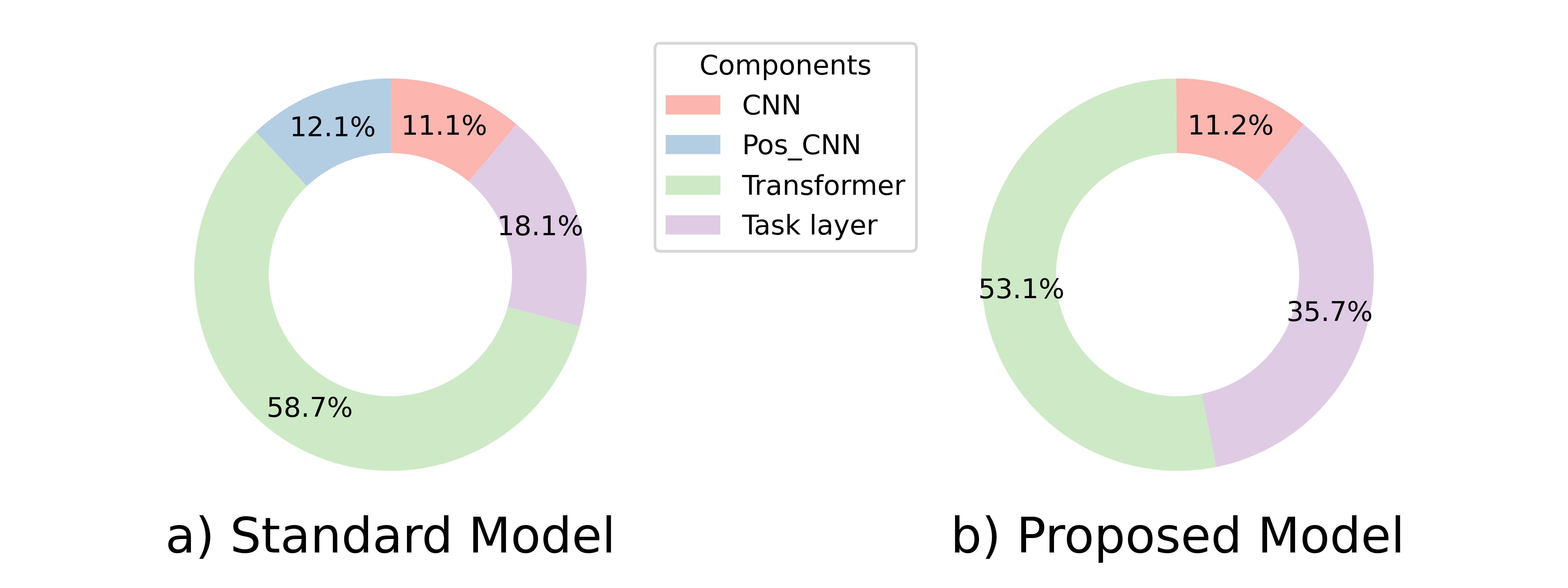}
\caption{Standard Architecture vs Proposed Architecture, where the size is measured after quantization. }
\label{fig:order}
\end{figure}

\vspace{0.1in}
\textbf{Model pruning:}
To find the best starter model for reaching the 10 MB goal, we first analyze the size of each component of the standard architecture, shown in Figure~\ref{fig:order}. We found that the positional convolution layer is 12.1\% of the model size and the CNN component is 11.1\% of the model size, while the task-specific layers only occupy 18.1\% of the model size. We hypothesized that simply removing the positional convolution layer and reducing the ratio of CNN parameters might be suitable for our specific use case. As pointed out in~\cite{fu2022improving}, the input data is clipped in such a way that the start of the interruption will always be at the beginning of the 5-second clip, so positional info is not important. This is discussed in more detail in Section~\ref{subsection:result}. After removing the positional convolution layer, we can see our model assigns more parameters for the transformer and task layers, which is expected to achieve better accuracy compared to the same size standard model. To shrink the model in both width and depth, we applied a reduction to the transformer layers.

\textbf{Weight sharing:} A common practice in deep learning is to build a model in full parameterization, where each layer is exclusive and independent so that each parameter is used only once in a forward pass. However, recent studies have demonstrated that the shared parameterization model~\cite{ge2022edgeformer}, where each layer is exploited more than once in a forward pass, could achieve better accuracy given the same parameterization budget. Motivated by this, we compared the shared parameterization model with a full parameterization model and searched for the best hyper-parameters. In the end, the best result was achieved by setting each neighboring three transformer layers to share the same weights. We also found a stepping effect in size versus layers because of this setting.

\subsection{Quantization}
Quantization has been proven as an effective way to reduce model size. It can be categorized into two types: Post Training Quantization (PTQ) and Quantization Aware Training (QAT). While QAT applies pseudo quantization/dequantization during training, PTQ simply maps the model weights from float32 to int8 (for example) after the training is done so the accuracy loss is more significant. However, we found that for our WavLM\_SI model, if we only quantize the non-CNN part of our model and enable per-channel quantization, there is almost no accuracy loss from PTQ.

\vspace{-0.2in}
\section{Experiments}
\label{experiment.section}

\subsection{Data statistics}
Our dataset consists of 40,068 audio clips of 5 seconds in length, each containing two channels, with the right channel storing the potential interrupter and the left channel merging everyone else's speech. As shown in Table~\ref{table:dataset}, we randomly split the dataset into train, validate, and holdout subsets, each representing 75\%, 19\%, and 6\% of the total size, respectively. The holdout test set contains only speakers that are never seen in other subsets to simulate evaluating the model in the real-world use case. We will release our dataset as part of a future challenge.

We used crowd-sourcing to label most of the dataset into four types: \textit{backchannel} (utterances that do not intend to interrupt, e.g. ``uh-huh", ``yeah"), \textit{failed interruption}, \textit{successful interruption}, and \textit{laughter}~\cite{fu2022improving}. Each data sample got labeled by 7 individuals and we chose 70\% as the agreement level, which means at least 5 of them have to agree. The holdout set is labeled by internal experts to ensure the best labeling quality for evaluation.

\vspace{-0.1in}
\begin{table}[htbp]
\caption{\label{table:dataset}Number of Clips and Labels}
\centering
  \begin{adjustbox}{width=250pt}
  \begin{threeparttable}
\begin{tabular}{c|c|c|c|c|c}
\hline
 & \textbf{Backchannel} & \textbf{Failed Interruption} & \textbf{Interruption} & \textbf{Laughter} &
 \textbf{Total} \\
\hline
\textbf{Train}  & 14,591 & 3,292 & 9,622 &  2,455  & 29,960 \\
\hline
\textbf{Validate}  & 3,693 & 889 & 2,478 &  623  & 7,683\\
\hline
\textbf{Holdout}  & 1,378 & 211 & 588 &  248 &  2,425\\
\hline
\textbf{All}  &  19,662 & 4,392 &  12,688 &   3,326 &  40,068\\
\hline
\end{tabular}
  \end{threeparttable}
  \end{adjustbox}
\end{table}
\vspace{-0.1in}

\subsection{Results}
\label{subsection:result}

\begin{figure}[hpbt]
\centering
\includegraphics[width=1.0\linewidth]{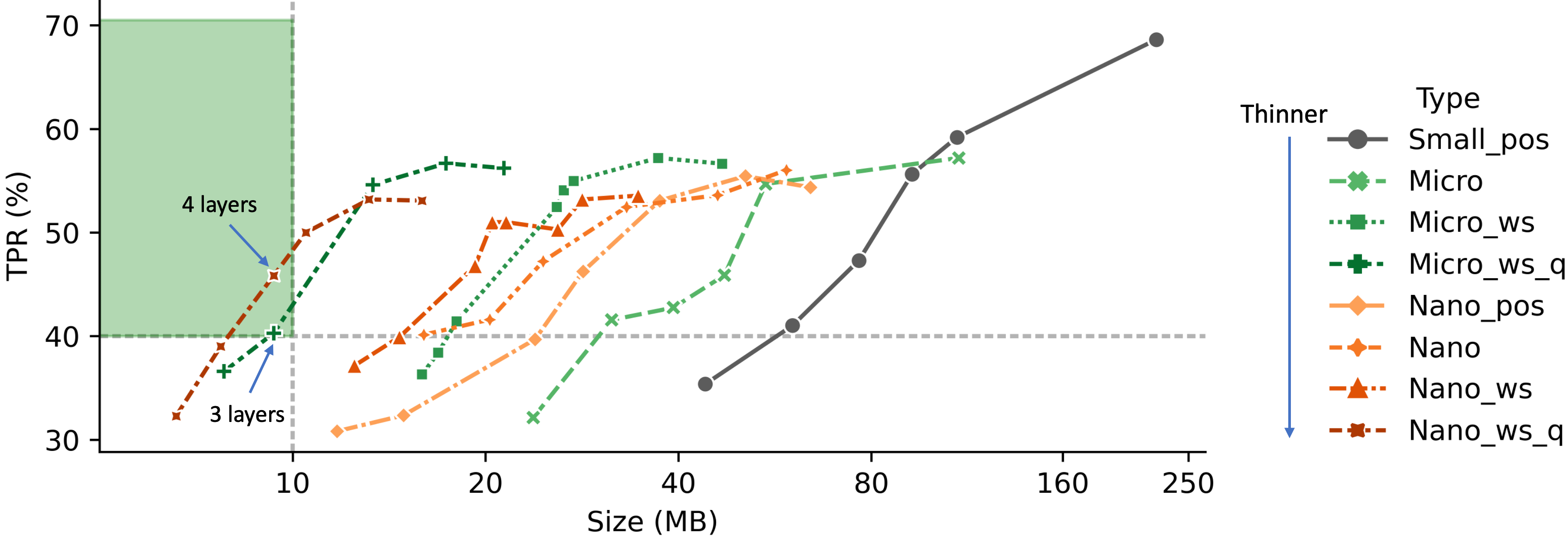}
\caption{True Positive Rate (TPR) is higher for thinner models. The green region is the desired performance range.}
\label{fig:sizeacc}
\end{figure}

We trained our models on 8 V100 GPUs for 70 epochs with a batch size of 16. All experiments are repeated 10 times and the average is reported. 
Table \ref{table:modelwidth} lists the TPR at 1\% FPR for the failed interruption class across different settings. Here we define thinness as the \textit{number of channels} for convolution layers and \textit{hidden size} for transformer layers, with \textit{Small, Micro, and Nano} referring to models that decrease in thinness. First, we can see that removing the positional convolution layer does not hurt the TPR by comparing \textit{Nano\_pos} and \textit{Nano}. Second, models that have shared weight perform similarly to their counterparts while hugely reducing the model size. Third, as shown in Figure~\ref{fig:sizeacc}, quantization is also effective in reducing the model size, and when done properly can have a minor or no accuracy drop. In the green region, both \textit{Nano\_ws\_q} with 4 layers and \textit{Micro\_ws\_q} with 3 layers meet our goal. We also tried Teacher-Student distillation, and applied layer drop during pre-training, but did not find improvement in the speech interruption detection task. 

We found that given a fixed model size, as shown in Figure~\ref{fig:sizeacc}, the model that is \textit{thinner} in the width dimension tends to have better accuracy. We also removed the positional convolution layer from our model since it has big kernels and makes up a large portion of the model size. We found no significant accuracy drop.

\vspace{-0.2in}
\begin{table}[htbp]
\caption{Complexity, RTF, and TPR by various parameters}
\centering
  \begin{adjustbox}{width=250pt}
  \begin{threeparttable}
\begin{tabular}{c|c|c|c|c|c|c}
\hline & \textbf{Small\_Pos\tnote{1}} & \textbf{Micro} & \textbf{Micro\_WS\tnote{2}} & \textbf{Nano\_Pos\tnote{1}} &
 \textbf{Nano} &
 \textbf{Nano\_WS\tnote{2}}\\
    \hline
    \textbf{\# Conv\_ch}  & 386 & 256 &  128 & 128 &128 & 128\\
    \hline\textbf{\# Trans\_hi}  & 576 & 384 &  384 & 288 & 288 & 288\\
    \hline
    \textbf{Param (M)}  & 54.0 & 24.4 &  8.1 & 13.6 &12.7 & 4.9\\
    \hline
    \textbf{MACs (G)} & 31.2 & 13.9 &  12.2 &7.9 & 6.4 & 7.6 \\
    \hline
    \textbf{RTF}  & 0.32 &  0.16 &   0.12 & 0.09 & 0.08 & 0.08\\
    \hline
    \textbf{TPR}  & 68.3\% & 57.2\% & 56.6\% &  54.4\% &56.0\%& 53.6\%\\
    \hline
    \end{tabular}

    \begin{tablenotes}
          \item [1] Pos: Models with positional convolution layers.
      \item [2] WS: Every three transformer layers share the same weights.
     \end{tablenotes}
     
  \end{threeparttable}
  \end{adjustbox}
\label{table:modelwidth}
\end{table}

Layer reduction and component removal can significantly reduce the model size with an acceptable accuracy drop. We tested different selections and ordering of transformer layer initialization on a vanilla 4-layer \textit{Nano} model, as shown in Table~\ref{table:layer_tpr}, and found that the position of the layers matters more than the layer orders, possibly due to residual connections within transformer layers. Previous research~\cite{lee2022fithubert, DBLP:conf/iclr/FanGJ20} has reported that \textit{Skip Layers} perform much better than other combinations, but we only observed a small increase, which may be because we applied fine-tuning after layer reduction. 

\begin{table}[htbp]
\caption{\label{table:layer_tpr}Layer Order vs True Positive Rate (TPR)}
\centering
  \begin{adjustbox}{width=250pt}
  \begin{threeparttable}
\begin{tabular}{c|c|c|c|c|c}
\hline
 \textbf{Layer Order} & 0,1,2,3 & 4,5,6,7 & 8,9,10,11 & 0,3,6,9 & 3,2,1,0 \\
\hline
\textbf{TPR}  & 45.88\% & 39.23\% & 28.44\% &  47.96\%  & 40.62\% \\
\hline
\end{tabular}
  \end{threeparttable}
  \end{adjustbox}
\end{table}

\vspace{-0.2in}
\subsection{Memory and Energy Analysis}

\begin{figure}[hpbt]
\centering
\includegraphics[width=1\linewidth]{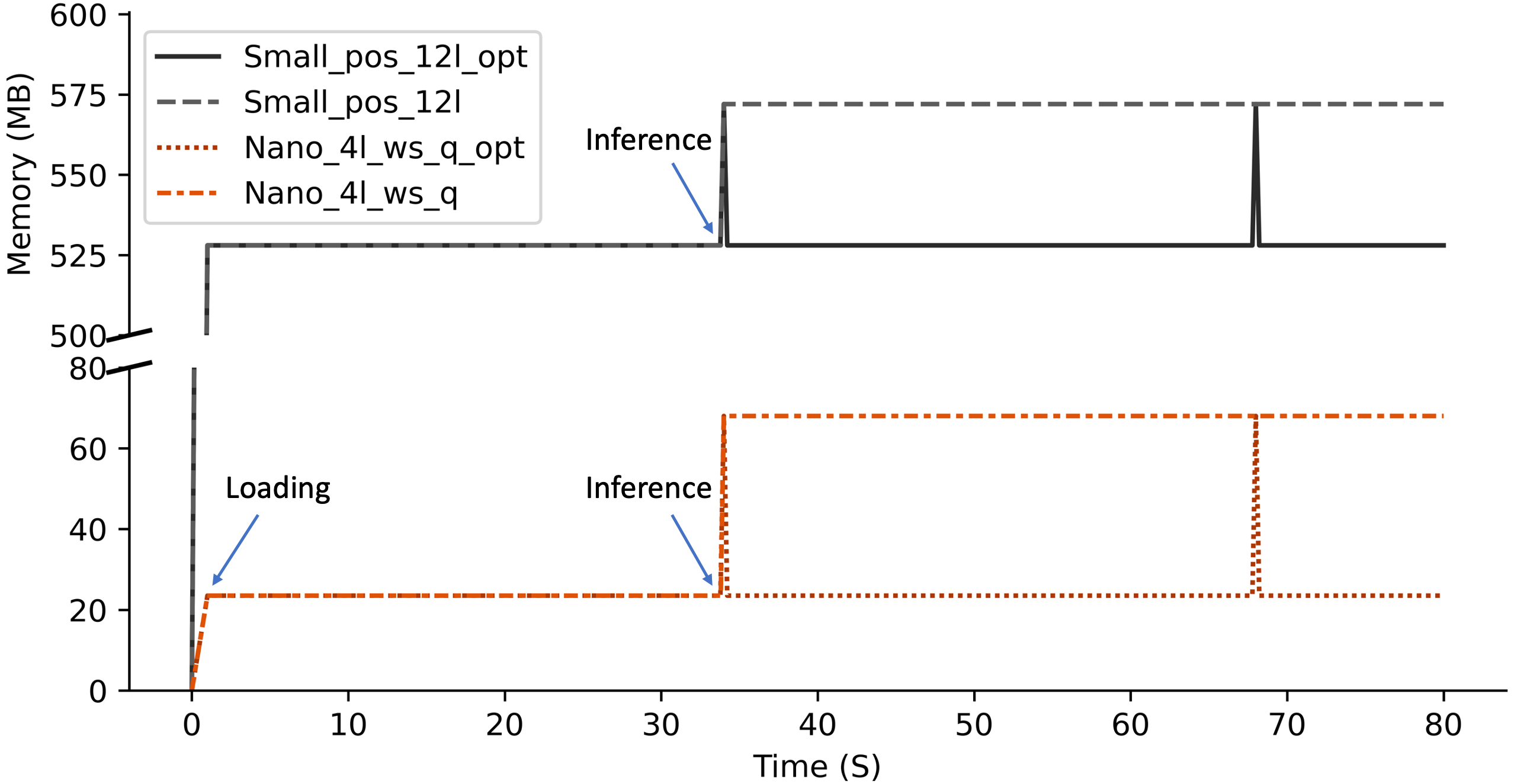}
\caption{Memory usage of the models with two ONNX parameter settings.}
\label{fig:mem}
\end{figure}

We simulated the memory usage by running our model with the ONNX runtime on an Intel Xeon E5-2673 CPU (2.40 GHz) with 6 GB of memory. We found that by turning off certain optimization options in the run time (\textit{arena\_mem}, \textit{mem\_pattern}, \textit{prepacking}, etc.) the amortized memory usage can be significantly reduced as shown in Figure~\ref{fig:mem} while also reducing energy usage. 

To analyze energy usage, we use inference time as a proxy for CPU usage. Using a  Voice Activity Detector (VAD)~\cite{braun2021training} we detected 57,450 overlaps of the human voice of more than 300 ms on 272 hours of meeting audio. Therefore on average, the inference should be triggered every 17 Seconds. Compared to naively doing inference every 5 seconds which is worse at capturing interruption clips, using a VAD to detect speech overlaps can save $\frac{17}{5}=3.4 \times$ inference calls. As VAD is a universal service that already exists in the client, the additional energy consumption is zero. The inference time was reduced from 1.6 seconds to 220 ms ($7.3 \times$), and the combined energry reduction is $25 \times$.

We measured the energy usage of our models by running a program that simulates the behavior of client deployment by doing an inference every 17 seconds and 5 seconds respectively for 100 times. After removing the baseline wattage of our machine (7.16W), the average power of our best model is 0.19W, and the average energy usage of our previous model is 4.84W. That makes a $25 \times$ reduction, which is the same as our estimation.

Considering 300 million feature users and assuming 4 hours of virtual conference time on weekdays, our techniques on model optimization and deployment savings are enough to support the electricity need of 92,711 people$^2$, which is the population of a small city; see Table~\ref{tab:power}.

\begin{table}
\caption{\label{tab:power}Estimated yearly power consumption of WavLM\_SI}
\centering
\begin{tabular}{lr} 
\hline 
Unoptimized feature per user	& 1.01 kWh \\
Optimized feature per user & 0.04 kWh \\
Unoptimized feature 300M users & 302 GWh \\
Optimized feature 300M users & 12 GWh \\
Savings 300M users & 290 GWh \\
Worldwide power usage per capita & 3,128 kWh \\
Savings 300M users in per capita usage	&  92,711  people \\
\hline
\end{tabular}
\end{table}

\subsection{Distribution Analysis}

\begin{figure}[hpbt]
\centering
\includegraphics[width=1.0\linewidth]{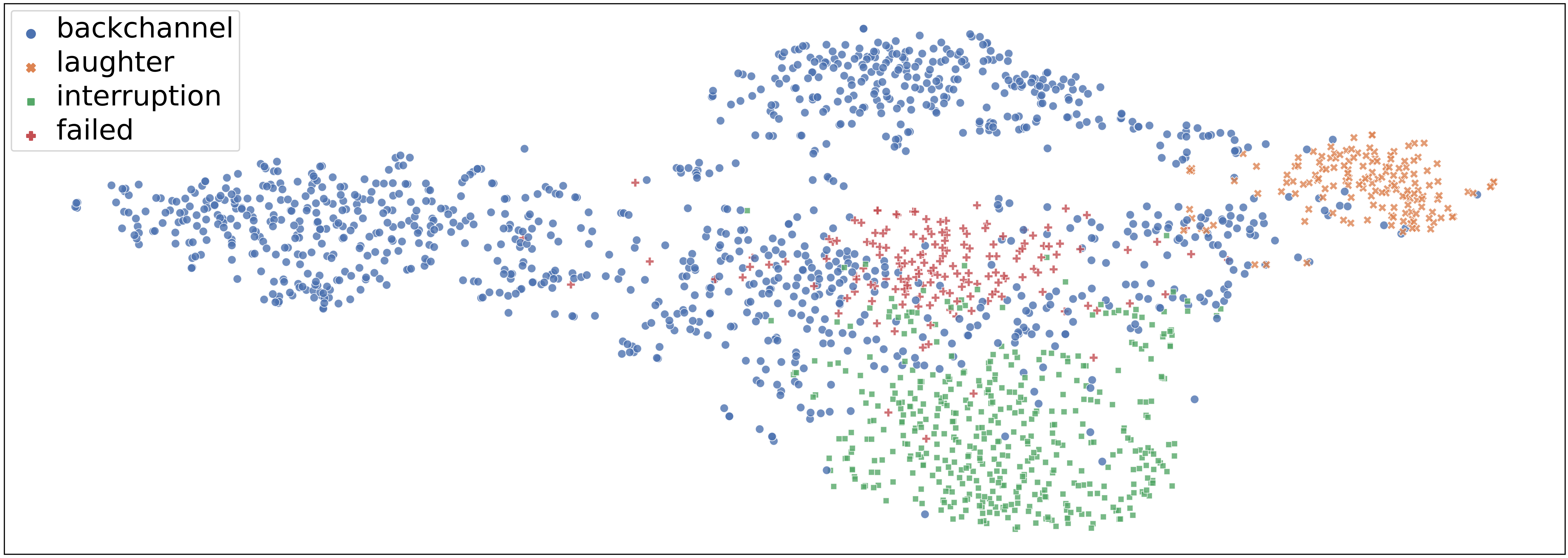}
\caption{T-SNE visualization of clip
embeddings.}
\label{fig:tsne}
\end{figure}

As shown in Figure~\ref{fig:tsne}, we plotted the T-SNE visualization for the model embedding of clips after the pooling layer. We found that the backchannel does not cluster into a tight space, which suggests it is hard to define a single type of backchannel. The backchannel is most likely to be confused with failed interruption, and in both cases, the active speaker does not give up the floor. Failed Interruption is very close in space to Successful  Interruption, but it can be linearly separated. Laughter is very different from other classes and can be easily separated.

\vspace{-0.2in}
\section{Conclusion}
\label{conclusion.section}

This paper investigates different techniques to reduce the WavLM\_SI model size and complexity to enable it to run on client devices and have less environmental impact. We successfully identified the CNN bottleneck and applied model pruning, weight sharing, and quantization to reduce the model size from 222.7 MB to 9.3 MB with acceptable accuracy loss, and further reduced the RTF from 0.32 to 0.04, which enables deploying the SSL model to the client. In future work, we will further improve the performance of the model by deploying the model to neural processing units (NPUs) and continue to improve the model's accuracy. 
\footnotetext[2]{The average worldwide electricity consumption per capita is 3,128 Kwh as estimated by the World Bank}

\bibliographystyle{IEEEbib}
\bibliography{failedspeech}

\end{document}